\pdfoutput=1

\documentclass[11pt]{article}

\usepackage[]{acl}

\usepackage{times}
\usepackage{latexsym}

\usepackage[T1]{fontenc}

\usepackage[utf8]{inputenc}

\usepackage{microtype}

\usepackage{amsmath}
\usepackage{amsfonts}
\usepackage{graphicx}
\usepackage{booktabs}
\usepackage{array}
\usepackage{pifont}
\usepackage{makecell}
\usepackage{tabularx}
\usepackage{xcolor}

\newcolumntype{C}[1]{>{\centering\let\newline\\\arraybackslash\hspace{0pt}}m{#1}}

\DeclareMathOperator*{\argmax}{argmax}

\title{RetICL: Sequential Retrieval of In-Context Examples with Reinforcement Learning}

\author{Alexander Scarlatos \and Andrew Lan \\
        University of Massachusetts Amherst\\
        \texttt{\{ajscarlatos,andrewlan\}@cs.umass.edu}}

\begin{document}

\maketitle

\begin{abstract}
Recent developments in large pre-trained language models have enabled unprecedented performance on a variety of downstream tasks. Achieving best performance with these models often leverages in-context learning, where a model performs a (possibly new) task given one or more examples. However, recent work has shown that the choice of examples can have a large impact on task performance and that finding an optimal set of examples is non-trivial. While there are many existing methods for selecting in-context examples, they generally score examples independently, ignoring the dependency between them and the order in which they are provided to the model. In this work, we propose Retrieval for In-Context Learning (RetICL), a learnable method for modeling and optimally selecting examples sequentially for in-context learning. We frame the problem of sequential example selection as a Markov decision process and train an example retriever using reinforcement learning. We evaluate RetICL on math word problem solving and scientific question answering tasks and show that it consistently outperforms or matches heuristic and learnable baselines. We also use case studies to show that RetICL implicitly learns representations of problem solving strategies.
\end{abstract}

\section{Introduction}

With the rising prominence of large pre-trained language models (LLMs), prior work has focused on how to best utilize them for various natural language tasks. One of the most popular methods for doing so is prompt tuning, which deals with carefully selecting the natural language prompt that maximizes model performance \cite{prompting}. While there are many approaches to prompt tuning, a very successful one is in-context learning (ICL) \cite{icl}. In ICL, examples of a new task that the LLM may not have been trained on before are included in the prompt, enabling it to leverage patterns in these examples in a few-shot way. However, the choice of which examples the LLM sees for a particular task can significantly affect the model's performance \cite{prompt-tuning}.

The primary goal of ICL example selection is to find examples that, when used in the prompt, elicit a desired response from an LLM. Common practice is to define a function to measure the quality of a set of examples as $\phi(x,e_1,\ldots,e_T)$, where $x$ is the input for the current task and $e_1,\ldots,e_T$ is a list of $T$ examples drawn from a corpus $\mathcal{C}$, and use $\phi$ to rank candidate examples. Most existing works assume that examples work independently of each other, i.e., $\phi(x,e_1,\ldots,e_T) = \prod_{t=1}^T\phi'(x,e_t)$. Thus, one can find the best set of examples by selecting the top $T$ examples in the corpus with the highest values of $\phi'(x,e_t)$, which is often the semantic similarity between $x$ and $e_t$. However, there is significant interplay between the roles of different examples in deciding the output of LLMs. Some tasks benefit from example diversity \cite{selective-annotation}, while others benefit from combining specific information across examples \cite{diverse-prompting}. In these cases, simply selecting the top-$T$ ranked examples may neglect ones that are ranked lower on their own but are useful in conjunction with other examples. Additionally, top-$T$ selection ignores the \textit{order} in which examples are provided as LLM input, which also has an impact on its output \cite{prompt-ordering}. See Section~\ref{sec:rw} for a more detailed discussion of related work on ICL example selection. 

\subsection{Contributions}

In this work, we propose RetICL (Retrieval for In-Context Learning), a fully learnable method that sequentially retrieves ICL examples by conditioning on both the current problem and examples that have already been selected. We frame the problem of sequential example selection as a Markov decision process (MDP) and train an example retriever model using reinforcement learning (RL). We construct the model using a recurrent architecture where hidden states act as latent representations of MDP states, and model the example ranking function using a bilinear transformation between the latent and corpus spaces, enabling efficient inference-time maximization of $\phi(x,e_1,\ldots,e_T)$. We also propose a novel \textit{confidence} reward function, which uses the perplexity of the \textit{generated} solution to help guide training. We validate RetICL on the math word problem (MWP) solving datasets TabMWP and GSM8K where it outperforms or matches both heuristic and learnable baselines. Additionally, to test RetICL's ability to generalize across domains, we validate it on the scientific question answering dataset QASC, where RetICL outperforms all baselines.
Finally, we qualitatively analyze RetICL's learned policies and find that RetICL is able to implicitly infer problem solving strategies while learning its ICL example selection policy. We intend to make our implementation publicly available.

\section{Related Work}
\label{sec:rw}

In ICL, it is common to either randomly select in-context examples \cite{icl, minerva} or use a hand-crafted set of examples \cite{mmlu, cot}. However, it is now well known that example selection and ordering can have a large impact on downstream text generation performance \cite{corr-fewshot, similarity, prompt-tuning, prompt-ordering}. There are many existing methods for in-context example selection that focus on different aspects of the problem. Several seek to maximize how much coverage a set of examples provides over the dataset and use diversity to measure coverage \cite{diverse-prompting, selective-annotation, pitis2023boosted}. Other methods consider semantic features of the examples \cite{similarity, fu2022complexity} or compare these features with outputs of the target LLM \cite{ids-prompting}. \citet{data-curation} use random prompting but with a curated corpus based on how well examples perform on a validation set, and \citet{icl-contrastive} use contrastive learning to retrieve examples that are likely to have similar labels to the target. 
While each of these methods tends to tackle a single aspect of example selection, we distinguish our method by combining several of these aspects, particularly considering groups of examples (including ordering) and using the LLM's outputs for a training signal.  
There are other works that use RL for ICL example selection \cite{tab-mwp,active-example-selection}, although they either do not include previously selected examples in the state or only use high-level features of the examples, while RetICL uses their exact textual content.

\section{Methodology}

\begin{figure*}[]
    \centering
    \includegraphics[width=\linewidth]{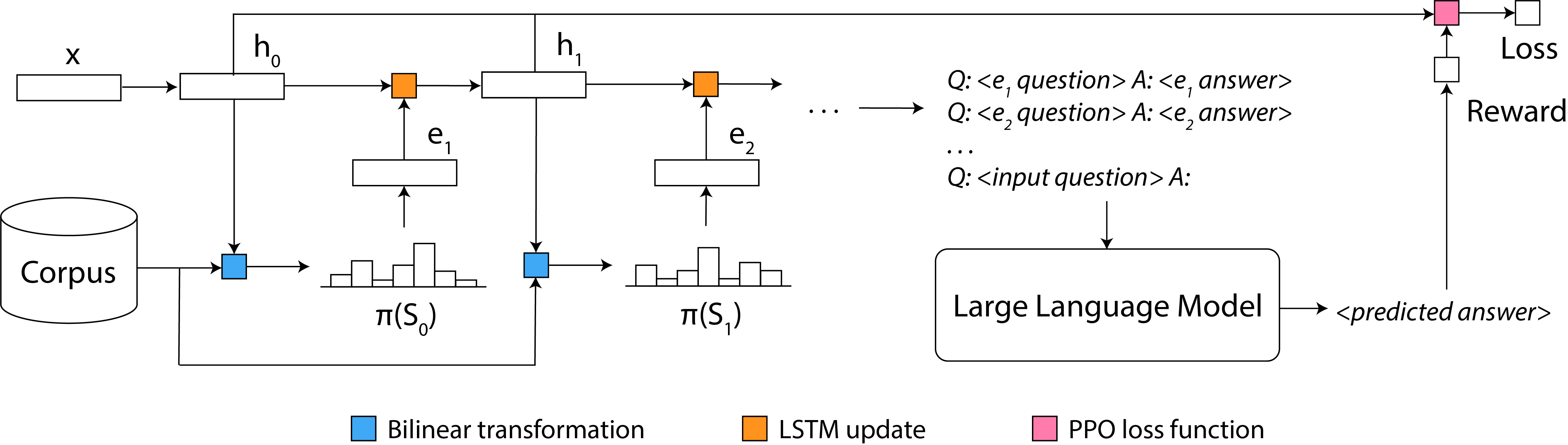}
    \vspace{-.5cm}
    \caption{RetICL overview for question answering. Each latent state is constructed from the problem and previously selected examples. The next example is selected using a bilinear transformation between the latent state and examples in the corpus. After all examples are selected, we query the LLM, obtain a reward, and update the policy.}
    \label{fig:arch}
\end{figure*}

In this section, we detail how we frame ICL example selection as an MDP, how our example retriever model works, and how we train it using RL. We show an overview of our methodology in Figure \ref{fig:arch}.

\subsection{MDP Formulation and Reward Function}

We can view ICL example selection as a sequential decision making problem, where we select examples one at a time in such a way that we maximize our chances of achieving some goal when the examples are used as context. Our goal is to maximize $r(\mathcal{M}(x,e_1,\ldots,e_T), y)$, where $\mathcal{M}(\cdot)$ returns the generated output of an LLM given a prompt, $y$ is the label corresponding to $x$, and $r$ is a task-specific function that returns how good the generated output is. We note that in this setup, the order in which examples are selected matters since the order in which they are provided to $\mathcal{M}$ must be defined. We also note that while in this work we set $T$ to a constant, it can also be dynamically set during the decision-making process, which we leave for future work. With this framing, we can naturally define an MDP where the state at time step $t$ corresponds to both $x$ and the first $t$ examples that have been selected, and the action space is the set of potential candidates to be the next example. Formally,
\begin{align*}
S_0 = x, \quad S_t = x, e_1, \ldots, e_t, \quad A_t = e_{t+1} \in \mathcal{C}.
\end{align*}

We now define the reward function for the MDP, which we break into two parts: a task-specific \textbf{goal} reward, and a supplementary \textbf{confidence} reward. We define the goal reward, $R^G$, simply as the output of $r$, as long as it can be formulated to return a scalar value. In settings with definitive correct and incorrect answers, it is natural for $r$ to be binary, where it returns 1 when the generated solution results in a correct answer and -1 when the generated solution results in an incorrect answer.
However, with chain-of-thought (CoT) prompting \cite{cot}, this reward function does not account for the reasoning strategy in the solution process that led to the final answer and thus cannot distinguish between sound and flawed logic.
To address this issue, we introduce the confidence reward, $R^C$, which we define as the inverse perplexity of the generated solution assigned by the LLM, normalized to the range $[-1,1]$. We hypothesize that when an LLM generates a correct solution with high probability (low perplexity), it is likely that the model ``knew'' how to solve the problem, rather than getting it correct by guessing or using unsound reasoning to arrive at a final answer. Additionally, we hypothesize that when an LLM generates an incorrect solution with high probability, it may have sound reasoning overall but contain a small error, such as an incorrect calculation. Recent works have also found that model confidence is predictive of downstream performance \cite{calibration, lmsknowwhattheyknow}.
We define the final reward function to be the average of $R^G$ and $R^C$ at the final time step and 0 at all prior time steps. We formally define our reward function as
\begin{align*}
    & \hat{y} = \mathcal{M}(x,e_1,\ldots,e_T),\\
    & R^G = r(\hat{y}, y) = 2\cdot \mathbb{I}[g(\hat{y}, y)] - 1,\\
    & R^C = 2\cdot p_{\mathcal{M}}(\hat{y}|x,e_1,\ldots,e_T)^{\frac{1}{|\hat{y}|}} - 1,\\
    & R_t = \begin{cases} 0.5 R^G + 0.5 R^C & \text{if } t = T\\ 0 & \text{otherwise}, \end{cases}
\end{align*}
where $\hat{y}$ is the generated solution, $g$ is a function that checks if two solutions have the same final answer, $\mathbb{I}$ is the indicator function, and $p_{\mathcal{M}}$ returns the probability assigned by the LLM.

\subsection{Retriever Model}

We now detail our model for example retrieval. At a high level, the model constructs a latent representation for each state $S_t$ in the MDP and uses this representation to construct the policy $\pi(S_t,e)$, which represents the probability of selecting $e$ to be the next example. After using the policy to select an example, we add it to the current sequence of examples, giving us the state $S_{t+1}$, and the process continues sequentially.

We use a long short-term memory (LSTM) model \cite{lstm} as the base model, where the hidden state $\mathbf{h}_t$ acts as the latent representation for $S_t$. We construct the initial hidden state of the LSTM, $\mathbf{h}_0$, using a vectorized embedding of the input $x$, and set the input of the LSTM at time step $t$ to be a vectorized embedding of the example $e_t$. In this work, we construct these vectorized embeddings using a pre-trained S-BERT model \cite{sbert} and additionally provide learnable soft prompts \cite{soft-prompts} to S-BERT to help align the embeddings with the current task. In our experiments, we found that fine-tuning the S-BERT parameters directly did not improve performance.

We produce the policy $\pi(S_t,e)$ by first producing an unnormalized \textit{activation} value for each example in the corpus, $\phi(S_t, e)$, and then using the softmax function to convert these activations into a probability distribution. We construct each $\phi(S_t, e)$ by performing a learnable bilinear transformation between $\mathbf{h}_t$ and the vectorized embedding of $e$. We choose to model $\phi$ using a bilinear transformation for two reasons. First, the bilinear transformation learns a mapping between the model's latent space and the example embedding space, enabling generalization to examples not seen during training and also adding some model interpretability, as we will show later in this paper. Second, the bilinear transformation enables efficient computation of the policy over a large corpus at inference time, which we describe in detail in Supplementary Material \ref{sec:mips}. We additionally use $\mathbf{h}_t$ to produce an estimate of the value function, $\hat{v}(S_t)$, which is required for variance reduction techniques when training policy gradient methods. Concretely, our model architecture is defined as
\begin{align*}
    & \mathbf{x} = \operatorname{S-BERT}(\mathbf{P}_x, x), \quad \mathbf{e} = \operatorname{S-BERT}(\mathbf{P}_e, e), \\
    & \mathbf{h}_0 = \tanh(\mathbf{W}_x \mathbf{x} + \mathbf{b}_x), \\
    & \mathbf{h}_{t>0} = \operatorname{LSTM}(\mathbf{h}_0; \mathbf{e}_1,\ldots,\mathbf{e}_t), \\
    & \hat{v}(S_t) = \mathbf{h}_{t}^T\mathbf{w}_v + b_v, \\
    & \phi(S_t, e) = \begin{cases}
        \mathbf{h}_{t}^T\mathbf{W}_a\mathbf{e} & \text{if } e \notin \{e_1,\ldots,e_t\} \\
        -\infty & \text{otherwise},
    \end{cases} \\
    & \pi(S_t, e) = \exp(\phi(S_t, e)) / \textstyle \sum_{e^\prime \in \mathcal{C}} \exp(\phi(S_t, e^\prime)),
\end{align*}
where $\mathbf{P}_x \in \mathbb{R}^{m \times d_e}$ and $\mathbf{P}_e \in \mathbb{R}^{m \times d_e}$ are learnable soft prompts, $\mathbf{W}_x \in \mathbb{R}^{d_h \times d_e}$ and $\mathbf{b}_x \in \mathbb{R}^{d_h}$ transform the input embedding space into the latent space, $\mathbf{w}_v \in \mathbb{R}^{d_h}$ and $b_v \in \mathbb{R}$ produce the value function estimate from the latent space, $\mathbf{W}_a \in \mathbb{R}^{d_h \times d_e}$ performs the bilinear transformation between the latent space and example embedding space, $m$ is the soft prompt length, $d_e$ is the dimension of the S-BERT text embedding vector, and $d_h$ is the size of the LSTM's hidden states. We set $\phi(S_t,e)$ to $-\infty$ when $e$ has already been selected to avoid selecting the same example multiple times, which is in line with existing methods.

\subsection{Training and Inference}

We train the retriever model using proximal policy optimization (PPO) \cite{ppo} with generalized advantage estimation (GAE) \cite{gae} as our advantage function. We use a reward discount of $\gamma=1$ since all episodes have fixed length and the reward is assigned only at the final time step. We train the value function estimator using mean squared error (MSE) with $R_T$ as the target at each time step and weigh the value function loss with a hyperparameter $c_{\text{VF}}$. We also encourage exploration by adding the negative entropy of the policy at each time step to the loss \cite{entropy}, where we additionally weigh the entropy by a hyperparameter $c_{\text{E}}$ and normalize by a factor of $\frac{1}{\log(|\mathcal{C}|)}$ to account for training with different corpus sizes.

At training time, we select a batch of problems from the dataset, encode their inputs with S-BERT, encode all examples in the corpus with S-BERT, and then construct a sequence of examples for each problem by sequentially sampling from the policy, i.e., $e_{t+1} \sim \pi(S_t, \cdot)$. When $T$ examples have been selected for each problem, we prompt the LLM with the examples and the current problem's input, calculate the reward from the LLM's generations, average the PPO loss, value function loss, and entropy loss over the batch, and backpropagate through our model. At inference time, we greedily select examples from the policy as $e_{t+1} = \argmax_{e \in \mathcal{C}}\pi(S_t, e)$, which empirically outperforms sampling in our experiments.

\section{Experiments}
\label{sec:expts}

In this section, we validate RetICL on MWP solving and scientific question answering tasks and quantitatively compare its performance against several baselines. We also perform an ablation study and examine the effects of adjusting several parameters in order to determine which aspects of the methodology work well and which ones need improvement in future work.

\subsection{Datasets}

We validate RetICL on two MWP datasets that contain detailed solution steps: \textbf{TabMWP} \cite{tab-mwp}, where solving each problem requires extracting and reasoning with information from tables, and \textbf{GSM8K} \cite{gsm8k}, where solving each problem requires multi-step mathematical reasoning and applying various arithmetic operations. In order to test whether RetICL generalizes beyond MWP solving, we also validate on \textbf{QASC} \cite{qasc}, which contains multiple choice science questions along with explanations for each answer, where answering each question requires logical deduction and leveraging world knowledge.

For all datasets, we use CoT prompting and evaluate correctness based on the final answer of the generated solution.
We note that the official TabMWP evaluation code has some issues with regular expressions that cause both false positives and false negatives when evaluating correctness on multiple choice problems. We instead use our own, fixed code to evaluate correctness on TabMWP. We provide additional information on each dataset and the prompts we use in Supplementary Material \ref{sec:datasets}.

\subsection{Experimental Settings}

For each dataset, we randomly select 5,000 problems for training and an additional 200 problems for the corpus, both from the original training set.
While it is possible to use a larger corpus, e.g., all remaining problems in the training set, we find that training on a smaller corpus results in higher accuracy. We randomly select 500 problems from the validation set to evaluate on after each epoch and save the model with the highest accuracy on this set. For validation and testing, we use the entire training set as the corpus.
We use OpenAI's \textit{code-davinci-002} Codex model \cite{codex} as the LLM for all experiments since it is free and works well on our task, and we leave experimentation on larger, more expensive models for future work.
We set the number of in-context examples to $T=2$ for all experiments to enable a fair comparison to other methods \cite{tab-mwp}. We note that while increasing $T$ tends to increase performance for heuristic methods, we find that RetICL's performance does not tend to increase with $T$ as expected. We provide details on the impact of $T$ in Supplementary Material \ref{sec:scaling-k} and leave further exploration of increasing $T$ for future work.
We provide additional hyperparameter settings and implementation details in Supplementary Material \ref{sec:hyperparameters}.

\subsection{Baselines}

We compare RetICL to the following baselines for in-context example selection, comprising both heuristic-based and learnable methods. We also estimate a performance upper bound by exhaustively checking all possible example combinations.

\noindent\textbf{Random}
With random selection, for each problem, we randomly sample $T$ unique examples from the corpus for the ICL prompt. We evaluate random selection on 3 random seeds and report the average accuracy across all 3 runs.

\noindent\textbf{kNN}
With kNN selection \cite{similarity}, for each problem, we select the $T$ examples with the most similar problem inputs from the corpus and use those for the ICL prompt, putting more similar examples later in the prompt due to recency bias \cite{prompt-tuning}. We evaluate similarity according to the Euclidean distance between the S-BERT embeddings of the problem inputs using the same pre-trained S-BERT model as RetICL.

\noindent\textbf{Complexity}
With complexity-based selection \cite{fu2022complexity}, for each problem, we randomly select $T$ examples that have the most \textit{complex} reasoning in the label for the ICL prompt. For TabMWP and GSM8K, we define complexity as the number of steps in the solution, and for QASC we define it as the number of words in the label.

\begin{table*}[tp]
    \centering
    \small
    \scalebox{1}{\begin{tabular}{|l|c|c|c|c|c|c|}
        \hline
        \textbf{Method} & \multicolumn{2}{c|}{\textbf{TabMWP}} & \multicolumn{2}{c|}{\textbf{GSM8K}} & \multicolumn{2}{c|}{\textbf{QASC}}\\
        & \textbf{Acc.} & \textbf{Ex.} & \textbf{Acc.} & \textbf{Ex.} & \textbf{Acc.} & \textbf{Ex.}\\
        \hline
        \hline
        Exhaustive & 98.30 & 37 & 97.95 & 47 & 98.49 & 36 \\
        \hline
        Random & 72.04 & 11,203 & 57.19 & 2,153 & 70.41 & 1,635 \\
        kNN \cite{similarity} & \textbf{88.95} & 10,003 & 59.74 & 1,883 & 61.99 & 964 \\
        Complexity \cite{fu2022complexity} & 63.80 & 281 & 54.66 & 13 & 74.19 & 3 \\
        PromptPG \cite{tab-mwp} & 73.43 & 7 & 56.94 & 8 & 73.65 & 2 \\
        LSTM Classifier & 77.21 & 13 & 64.82 & 4 & 69.65 & 8 \\
        RetICL & 88.58 & 407 & \textbf{66.11} & 97 & \textbf{76.13} & 135 \\
        \hline
    \end{tabular}}
    \vspace{-.2cm}
    \caption{Test set accuracy and examples selected for RetICL and baselines.}
    \label{tab:results}
\end{table*}

\noindent\textbf{PromptPG}
With PromptPG \cite{tab-mwp}, for each problem, a learned scoring function is evaluated on each individual example in the corpus, and the top $T$ scoring examples are selected for the ICL prompt.
While PromptPG also uses RL to learn its example selection policy, there are many key differences between their method and ours: they do not include previously selected examples in the state, they do not use a confidence-based reward, and they use a much simpler RL framework.
We evaluate PromptPG's performance by running their code with modifications to match our prompting style and use our fixed evaluation code.

\noindent\textbf{LSTM Classifier}
In order to determine the effectiveness of RetICL's training pipeline, we train a model with a similar architecture but in a supervised manner. Specifically, using the same input format as RetICL, we train an LSTM classifier to predict, at each time step, if the prompt will result in a correct or incorrect response from an LLM. We train on 20 randomly sampled prompts for each sample in the training set with the same training set and corpus as RetICL. At inference time, we greedily select examples that result in the highest predicted likelihood of getting a correct response. We note that this setup is equivalent to estimating the $Q$-function of a random policy, and provide further details in Supplementary Material \ref{sec:lstm-classifier}.

\noindent\textbf{Exhaustive}
With exhaustive evaluation, for each problem, we construct a one-shot ICL prompt for each example in the corpus, and consider the current problem to be solved if a correct solution is generated from any of the prompts. We use one-shot prompts instead of few-shot prompts to reduce the search space. Additionally, we restrict the corpus size to 100 and only evaluate on the pre-defined 1,000-sample subset of the test set for TabMWP to reduce computation time.

\subsection{Results}

Table \ref{tab:results} shows the performance of RetICL and baselines on all datasets, with \textbf{Acc.}\ being problem solving accuracy and \textbf{Ex.}\ being the number of unique examples used for each method. We see that RetICL performs the best among non-exhaustive methods on all datasets, beating most baselines by a large margin, with the exception of TabMWP where kNN performs slightly better. We note that while the relative performance of baselines seems to depend on the dataset, RetICL performs well across datasets, showing that our methodology is more generalizable across tasks.

We note that kNN likely performs well on TabMWP due to the presence of many problems in the dataset with very high similarity. Many problems have exactly the same question text other than a few numbers or names changed, making it easy for the LLM to generate a correct solution given a highly similar example. On the contrary, GSM8K does not tend to contain problems that are almost identical, which makes kNN ineffective since problems with high textual similarity may not have similar solution strategies. Additionally, kNN performs worse than Random on QASC since we find that using examples that are highly similar to the current question can cause the LLM to copy answers from the examples and ignore the details of the current question. Furthermore, we see that Complexity is a poor heuristic for TabMWP since it does not account for problem similarity, and also for GSM8K since it uses some examples with overly abstract reasoning that appear to confuse the model.

Perhaps surprisingly, we see that PromptPG is only slightly better than Random on TabMWP and performs on par with Random on GSM8K. And while PromptPG is better than Random and kNN on QASC, we observe that it selects the same examples for each question, thus lacking example diversity. While the results on TabMWP contradict the trends reported in \cite{tab-mwp}, we believe the discrepancy is mostly due to using the fixed evaluation code. We believe that PromptPG's relatively low performance also highlights the challenges of solving the ICL example selection problem using RL; in practice, many training tricks are necessary to achieve high performance.

We see that the LSTM Classifier performs relatively well on GSM8K but not on other datasets. We believe the reason is that TabMWP and QASC require more targeted example selection strategies, which are difficult to extract from random example prompts due to sparse coverage of the exponentially large action space. This result highlights the benefit of on-policy learning for example selection.
We also see that the Exhaustive method achieves almost perfect accuracy on all datasets. We find this surprising, especially due to the fact that Exhaustive only uses a single ICL example and has access to a smaller corpus. This result implies that there is significant room for growth in ICL example selection methods and also implies that one-shot ICL has the potential to be extremely powerful as long as the example corpus is informative, even for challenging text generation tasks.

\begin{figure}[tp]
    \centering
    \includegraphics[width=.9\linewidth]{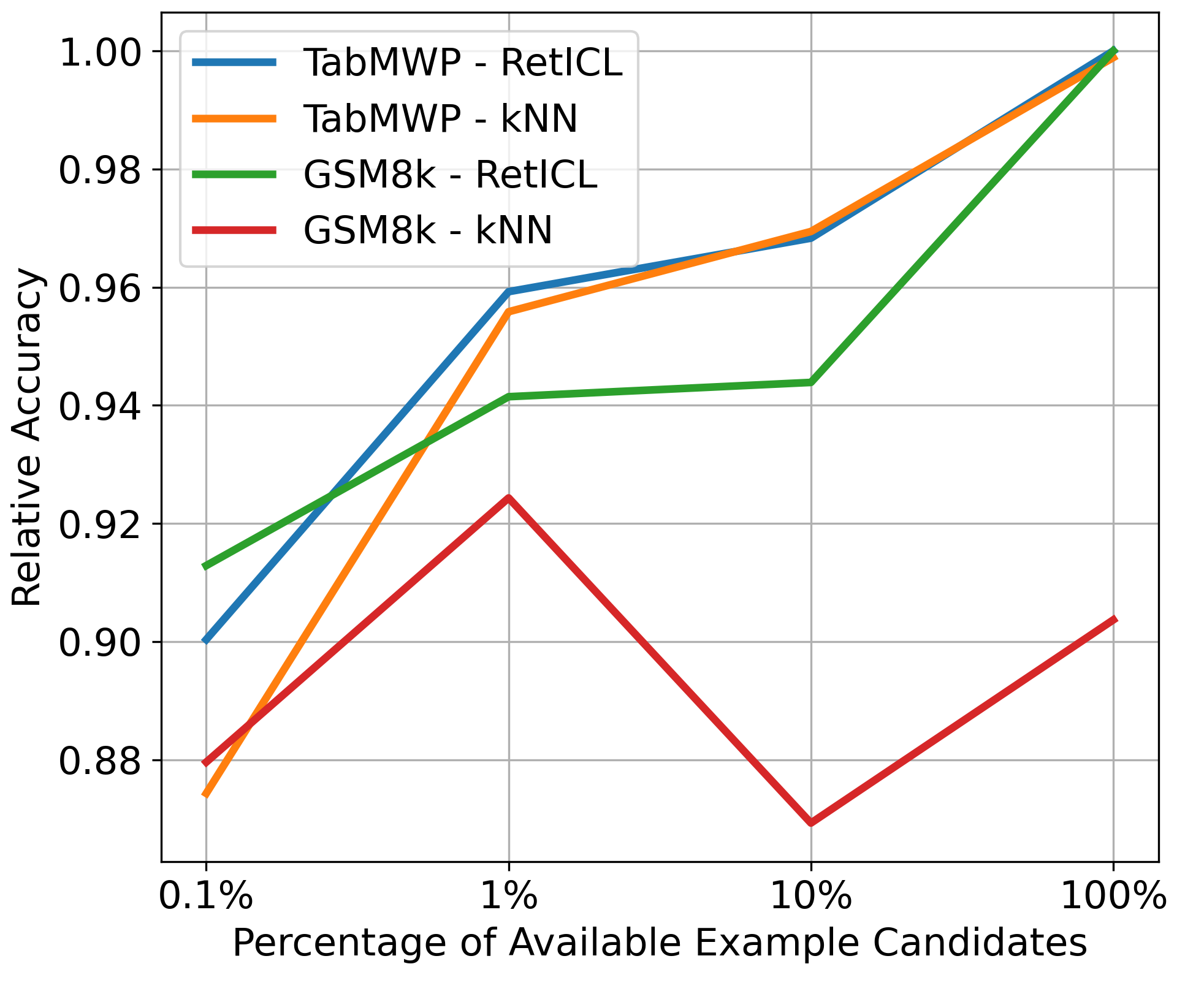}
    \vspace{-.2cm}
    \caption{Change in relative accuracy as number of available example candidates increases.}
    \label{fig:corpus_size}
\end{figure}

In order to determine how well RetICL can generalize to low-resource settings, we examine the effect of reducing the number of available examples at test time. We evaluate RetICL and kNN on TabMWP and GSM8K where we use 0.1\%, 1\%, 10\%, and 100\% of all examples as the corpus and show our results in Figure \ref{fig:corpus_size}. For a clearer visualization,
we show the \emph{relative accuracy} of each method compared to RetICL's accuracy when the full corpus is available. For TabMWP, we evaluate on the pre-defined 1000-sample subset of the test set. We see that while performance tends to decrease with less available examples, RetICL still retains at least 90\% of its performance with only 0.1\% of examples available, suggesting that it is still effective in low-resource settings. 
Additionally, we see that while RetICL and kNN perform similarly at most corpus sizes, RetICL performs better at 0.1\%, likely because there are not enough highly similar examples for kNN to leverage. Finally, we note that the irregular trend for GSM8K with kNN implies that semantic similarity of examples is not correlated with problem solving accuracy on this dataset, suggesting that while kNN is very effective for some tasks it is not as robust as RetICL.

\subsection{Ablation study}

We now examine the impact of various modeling and algorithmic choices via an ablation study. We train on 1,000 problems instead of 5,000 ($\textbf{T}_{\textbf{1k}}$), we no longer use the confidence reward, $R^C$, and instead only use the goal reward, $R^G$ (\textbf{Conf. Rew.}), we no longer condition on previously selected examples by removing the LSTM architecture and instead set the latent state for all time steps to be $\mathbf{h}_0$ (\textbf{LSTM}), we no longer include an entropy term in the loss function (\textbf{Ent.}), we no longer train with PPO and instead use REINFORCE (\textbf{R}) and REINFORCE with Baseline (\textbf{RwB}), we no longer provide learnable soft prompts to the S-BERT encoder (\textbf{SP}), and we vary the size of the corpus at train time, using a corpus with 20 problems ($\textbf{TC}_{\textbf{20}}$) from the training set and all remaining problems ($\textbf{TC}_{\textbf{all}}$) from the training set. Additionally, we use $\text{T}_{\text{1k}}$ for all ablations for fast experimentation, and apply the SP ablation for the TC ablations since otherwise the partial gradients over S-BERT parameters will cause memory issues for $\text{TC}_{\text{all}}$.

Table \ref{tab:ablation} shows the results of the ablation study, which we run on both TabMWP and GSM8K.
For TabMWP, we evaluate on the pre-defined 1,000-sample subset of the test set.
We see that removing the confidence reward and LSTM architecture, which are our key contributions, both negatively impact accuracy, implying that these techniques have a positive impact on ICL example selection. We also see that removing the entropy term has a significant negative impact on accuracy and example diversity, implying that this term is necessary for training the model. Furthermore, we see that REINFORCE with Baseline is similar or slightly worse than PPO, while REINFORCE is significantly worse. Surprisingly, we see that removing soft prompts slightly increases accuracy, although significantly hurts example diversity on GSM8K, due to training instability. We hypothesize that soft prompts may make it harder to find an optimal policy due to increased model complexity, and plan on finding better ways to fine-tune S-BERT in future work. Finally, we see that using much smaller and much larger example corpora at train time both negatively impact accuracy, showing that corpus size is an important hyperparameter and confirming similar results in \cite{tab-mwp}.

\begin{table}[tp]
    \centering
    \small
    \begin{tabular}{|l|c|c|c|c|}
        \hline
        \textbf{Ablation} & \multicolumn{2}{c|}{\textbf{TabMWP}} & \multicolumn{2}{c|}{\textbf{GSM8K}}\\
        & \textbf{Acc.} & \textbf{Ex.} & \textbf{Acc.} & \textbf{Ex.}\\
        \hline
        \hline
        None & 88.30 & 226 & 66.11 & 97 \\
        $\text{T}_{\text{1k}}$ & 87.30 & 159 & 65.96 & 34 \\
        $\text{T}_{\text{1k}}$, Conf. Rew. & 86.00 & 84 & 64.67 & 20 \\
        $\text{T}_{\text{1k}}$, LSTM & 85.10 & 120 & 63.91 & 38 \\
        $\text{T}_{\text{1k}}$, Ent. & 79.90 & 15 & 62.77 & 6 \\
        $\text{T}_{\text{1k}}$, R & 74.90 & 6 & 61.94 & 14 \\
        $\text{T}_{\text{1k}}$, RwB & 85.30 & 104 & 66.19 & 5 \\
        $\text{T}_{\text{1k}}$, SP & 88.40 & 163 & 66.26 & 3 \\
        $\text{T}_{\text{1k}}$, SP, $\text{TC}_{20}$ & 84.50 & 76 & 61.87 & 58 \\
        $\text{T}_{\text{1k}}$, SP, $\text{TC}_{\text{all}}$ & 85.50 & 122 & 65.88 & 58 \\
        \hline
    \end{tabular}
    \vspace{-.2cm}
    \caption{Ablation results on TabMWP and GSM8K.}
    \label{tab:ablation}
\end{table}

\section{Qualitative Analysis}

We now present several qualitative analyses to interpret RetICL's learned example selection policy. Our goal is to determine what features RetICL focuses on in individual examples and what strategy RetICL uses to select examples sequentially. We investigate these strategies by first visualizing learned latent example embeddings and then  analyzing trends in per-problem example selections.

\subsection{Latent Space Analysis}

In order to identify features in the selected examples that are being emphasized by RetICL, we perform a visual analysis of the example embeddings in the model's latent space. Specifically, we transform each example embedding $\mathbf{e}$ into the model's latent space using the right half of the bilinear term in $\phi$, i.e., $\mathbf{W}_a\mathbf{e}$. We note that since maximizing $\phi(S_t,e)$ is equivalent to maximizing the inner product $\langle \mathbf{h}_t, \mathbf{W}_a\mathbf{e} \rangle$, the most likely example to be selected is the one where $\mathbf{W}_a\mathbf{e}$ is closest to $\mathbf{h}_t$ in the latent space.\footnote{Maximum inner product and minimum distance are equivalent in our case since $\mathbf{e}$ is normalized.} Therefore, analyzing local regions in the example embedding space reveals how RetICL's learned policy ranks examples. 

For both TabMWP and GSM8K, we randomly select 1,000 examples from the corpus and then apply t-SNE \cite{tsne} to reduce their embeddings to 2 dimensions for visualization. Additionally, for the same sets of examples, we also visualize their pre-trained S-BERT embeddings in the same way in order to demonstrate how inter-example similarities change after RetICL training. We summarize our findings here and show the visualizations with a more detailed analysis in Supplementary Material \ref{sec:latent-space}.

For TabMWP, the S-BERT embeddings are clustered based on problem template, since problems tend to fall into distinct structural and semantic categories. While RetICL mostly retains these clusters in the latent space, it also merges together some clusters of problems that can be solved with similar reasoning strategies, such as finding the largest or smallest value in a set. For GSM8K, the clusters are much less clear since problems cannot be easily placed into categories. However, while the S-BERT embeddings are generally grouped by problem topic, RetICL also groups them by the number of steps in a solution. This result implies that RetICL has learned that solution length is an important feature of ICL examples, and confirms findings from prior work that solution complexity impacts problem solving in LLMs \cite{fu2022complexity}. These findings suggest that RetICL can identify meaningful features, often related to solution strategy, that indicate an example's utility in ICL.

\subsection{Per-Problem Example Selection}

We now examine example selections at the per-problem level in order to gain further insight into RetICL's learned example selection policy.

Table \ref{tab:example_gsm8k_pos} in Supplementary Material \ref{sec:examples} shows the in-context examples selected to help solve a representative problem from the GSM8K dataset. We see that RetICL tends to select examples that share some unique high-level features with the current problem, such as subtracting from a total value, adding up values over some period of time, or defining variables to be proportional to other variables. We note that each problem in the dataset exhibits several such features, so RetICL has to implicitly decide which features are important to the current problem and identify examples with those features. We also see that RetICL tends to select examples with solutions that are relatively long and have numerous reasoning steps. Problem solving errors can be divided into several categories. First, the LLM can exhibit misconceptions when it lacks an example to provide context, such as misinterpreting the meaning of a ``discount'' when not explicitly instructed. Second, the LLM can try to follow the examples too closely and use reasoning that does not necessarily apply to the current problem. These errors indicate that RetICL's policy can be improved by selecting based on a broader and more targeted set of features. However, many incorrect solutions are caused by simple arithmetic errors or switching the roles of variables in the problem. We believe these errors are due to limitations of the LLM and are a likely source of noise in the training signal, making it harder to find an optimal policy. Such errors can be fixed by using self-consistency \cite{self-consistency} or external computation engines \cite{wolfram}, which we leave for future work.

Table \ref{tab:example_tabmwp_pos} shows in-context examples selected for a problem from the TabMWP dataset. RetICL's selections tend to follow a surprising pattern: the first example is seemingly unrelated to the current problem, while the second example has similar reasoning steps to the current problem. This strategy has several implications. First, it suggests that RetICL can infer reasoning steps from the current problem and select examples based on this information. Second, it suggests that diversity of examples may help prevent overfitting to particular features in the examples.
Problem solving errors tend to be caused by either RetICL selecting an unrelated second example or the second example being similar to the current problem but requiring a slightly different solution strategy. Therefore, RetICL can benefit from policy consistency regularizations and improvements to the retriever model to learn more accurate solution strategy representations.

\section{Conclusions and Future Work}

In this work, we proposed RetICL, a learnable method for sequential in-context example selection that, unlike existing methods that select all examples independently, takes previously selected examples into account. We framed the problem of sequential example selection as a Markov decision process and developed a novel reward function and example retriever model. We demonstrated that RetICL learns effective strategies for example selection and consistently performs well across math word problem and question answering tasks. There are many avenues for future work. First, we can explore RetICL's effectiveness on a wider array of tasks, including ones with open-ended goals. Second, we can explore other architectural modifications that could further improve the retriever model, such as using a Transformer instead of an LSTM. Third, since we used a fixed number of examples, we can extend RetICL to let it learn how many examples are needed. Fourth, we can explore whether RetICL can be applied to real-world educational settings, e.g., selecting worked examples to help students solve practice problems.

\newpage

\section*{Limitations}

We note that there are several practical limitations to our method. First, we note that RetICL can be expensive and time-consuming to train, with each of our main training runs requiring up to 250,000 LLM inferences. This high number of inferences makes training on paid models prohibitively expensive; for example, it could cost up to approximately \$2,500 to train on OpenAI's \textit{text-davinci} models. Additionally, newer OpenAI models, such as \textit{gpt-3.5-turbo} and \textit{gpt-4}, do not return likelihood information on generated text, making the confidence reward impossible to calculate for these models. We finally note that RetICL's performance not increasing with the number of examples in the prompt may limit its use in practical settings.

\section*{Ethical Considerations}

We first note that the high number of inferences required to train RetICL give the method an outsized cost in terms of energy usage; however, we note that the method has a relatively low cost at inference time given its relatively low number of parameters and potential for optimization with MIPS. Additionally, we note that because RetICL uses a black-box LLM reward signal, its example selections are not guaranteed to be interpretable by humans. Finally, because we only experiment with question answering settings, we did not perform any analysis of bias in RetICL's selections. However, it is possible that RetICL could reflect biases in the LLM it is being trained on. As such, we recommend an analysis of bias in future works that use RetICL in sensitive settings such as student-facing educational tools.

\newpage

\bibliography{anthology,references}

\newpage

\appendix

\section{Hyperparameters and Implementation Details}
\label{sec:hyperparameters}

We implement the retriever model and RL algorithms in PyTorch. We note that our PPO implementation is slightly simpler than the standard implementation in that it does not use an inner training loop and instead takes a single training step on each batch sampled from the policy. We also note that we used GitHub Copilot to assist with minimal code-writing.  We encode problem inputs and examples using the \textit{all-distilroberta-v1} pre-trained S-BERT model with the \textit{sentence-transformers} library \cite{sbert}, take the normalized mean-pooled final layer outputs as the embeddings, and use a soft prompt length of 20. With Codex, we use greedy decoding and set the maximum number of generated tokens to 450/400/150 for TabMWP/GSM8K/QASC, respectively. We set the LSTM's hidden size to 800, PPO's $\epsilon$ to 0.1, GAE's $\lambda$ to 0.9, and $c_{\text{VF}}$ to 0.5. We set $c_{\text{E}}$ to 0.05 for TabMWP and $c_{\text{E}}$ to 0.1 for GSM8K, where different values are necessary since we find that our method performs differently across datasets and that $c_{\text{E}}$ has a large impact on training stability. We use orthogonal initialization \cite{orthogonal} for all weight parameters, initialize all bias parameters to 0, and initialize soft prompts using a standard normal distribution. We train using the AdamW optimizer for 50 epochs with a learning rate of 0.001, a weight decay of 0.01, and a batch size of 20. We additionally apply gradient norm clipping on all parameters using a value of 2, which we find is critical to avoid spikes in training losses. We provide the values we experimented with in Table \ref{tab:hyperparameters}, where we selected values based on preliminary experiments aiming to optimize both accuracy and clock time. All final experiments were run on a \textit{Lambda Vector} workstation with \textit{NVIDIA RTX A6000} GPUs. With a single thread, training a RetICL model for TabMWP with 5,000 training samples and 500 validation samples for 50 epochs takes approximately 44 hours, and inference for 1,000 TabMWP samples takes approximately 10 minutes. We note that we decrease runtime by using a cache to avoid repeated LLM inferences at both train and test time, batching requests to OpenAI, and using an adaptive backoff for waiting in between OpenAI API calls in order to maximize throughput given the API rate limit. Finally, we note that a RetICL model with the above hyperparameters has approximately 5.5 million parameters.

We use \textit{scikit-learn} for t-SNE and \textit{matplotlib} for visualizing data. All software used in this work is either open source or does not specify a license. To the best of our knowledge, we are consistent with the terms and intended use of all software and services, particularly OpenAI.

\begin{table}[h]
    \centering
    \begin{tabular}{|l|l|}
        \hline
        \textbf{Hyperparameter} & \textbf{Values Tried} \\
        \hline
        \hline
        Training size & 1,000, \textbf{5,000}\\
        Corpus size & 20, \textbf{200}, 500, 1,000, all\\
        Soft prompt length & 5, 10, \textbf{20}, 40\\
        LSTM hidden size & 100, 200, 400, \textbf{800}, 1600\\
        $\epsilon$ & \textbf{0.1}, 0.2\\
        $\lambda$ & \textbf{0.9}\\
        $c_{\text{VF}}$ & \textbf{0.5}, 1.0\\
        $c_{\text{E}}$ & 0.01, \textbf{0.05}, \textbf{0.1}, 0.2, 0.5\\
        Epochs & \textbf{50}, 100, 250\\
        Learning rate & 0.0001, 0.0005, \textbf{0.001}\\
        Weight decay & 0.001, \textbf{0.01}\\
        Batch size & 5, 10, \textbf{20}, 40, 64\\
        Gradient clipping & 0.5, 1.0, \textbf{2.0}\\
        \hline
    \end{tabular}
    \caption{Hyperparameter search space, where selected values are bold.}
    \label{tab:hyperparameters}
\end{table}

\section{Dataset Details}
\label{sec:datasets}

TabMWP has a pre-defined train/validation/test split of 23,059/7,686/7,686, GSM8K has a pre-defined train/test split of 7,473/1,319, and QASC has a pre-defined train/validation/test split of 8,134/926/920. We reserve 1,000 random problems from GSM8K's train set for validation. Because QASC's test set does not have labels, we instead use the validation set for testing and reserve 1,000 random problems from the train set for validation. All datasets are exclusively in English. TabMWP uses the CC BY-NC-SA license, GSM8K uses the MIT license, and QASC uses the Apache 2.0 license. We show our prompt templates, which we use for both S-BERT and the LLM, in Table \ref{tab:prompts}.

\begin{table*}[]
    \small
    \centering
    \begin{tabularx}{\linewidth}{|l|X|X|}
        \hline
        \textbf{Dataset} & \textbf{Template} & \textbf{Example} \\
        \hline
        \hline
        TabMWP & Table: [table]\newline Problem: [problem statement]\newline Solution: [solution steps]\newline Final Answer: [final answer] & Table: [TITLE]: Siblings\newline Number of siblings | Frequency\newline 0 | 14\newline 1 | 3\newline 2 | 6\newline 3 | 8\newline 4 | 0\newline Problem: The students in Mr. Boyer's class recorded the number of siblings that each has. How many students have at least 2 siblings?\newline Solution: Find the rows for 2, 3, and 4 siblings. Add the frequencies for these rows.\textbackslash n\textbackslash nAdd:\textbackslash n\textbackslash n6 + 8 + 0 = 14\textbackslash n\textbackslash n14 students have at least 2 siblings.\newline Final Answer: 14 \\
        \hline
        GSM8K & Problem: [problem statement]\newline Solution: [solution steps]\newline Final Answer: [final answer] & Problem: Marcos has to get across a 5 mile lake in his speedboat in 10 minutes so he can make it to work on time. How fast does he need to go in miles per hour to make it?\newline Solution: 10 minutes for 5 miles means 10 minutes / 5 miles = 2 minutes/mile\newline 1 hour is 60 minutes so 60 minutes/hour / 2 minutes/mile = 30 miles/hour\newline Final Answer: 30 \\
        \hline
        QASC & Question: [question] [options] \newline Solution: [fact 1] [ fact 2] [combined fact]\newline Final Answer: [final answer] & Question: Mussels have what? (A) seaweed (B) arms (C) Energy (D) a shell (E) warmth (F) bacteria (G) Length (H) legs\newline Solution: Most mollusks have shells. Mussels are bivalve mollusks. Mussels have shells.\newline Final Answer: a shell \\
        \hline
    \end{tabularx}
    \caption{Prompt template for a single in-context example and a formatted example for each dataset. We use the \textbackslash n\textbackslash n solution step delimiter for TabMWP to match the style in the original paper.}
    \label{tab:prompts}
\end{table*}

\section{Inference-Time Optimization}
\label{sec:mips}

We note that our formulation for $\phi$ enables efficient retrieval of the top-ranking example at each time step via maximum inner-product search (MIPS). We first note that $\phi(S_t,e)$ is maximized by finding the example $e$ that maximizes the inner product $\langle \mathbf{h}_t, \mathbf{W}_a\mathbf{e} \rangle$. We can leverage this information by first pre-computing $\mathbf{W}_a\mathbf{e}$ for each example in the corpus and constructing a MIPS index over these vectors, using a library such as \textit{faiss} \cite{faiss}. At inference time, we can now leverage algorithms that perform approximate MIPS in sublinear time, i.e., maximize $\phi(S_t,e)$ without evaluating $\mathbf{h}_t^T\mathbf{W}_a\mathbf{e}$ for each example in the corpus. We note that we do not use MIPS in this work since the corpora we experiment on are sufficiently small such that evaluating $\mathbf{h}_t^T\mathbf{W}_a\mathbf{e}$ for each example is relatively inexpensive. However, we expect that significant computational time can be saved with MIPS when evaluating on corpora at much larger scales.

\section{LSTM Classifier Details}
\label{sec:lstm-classifier}

For our LSTM Classifier baseline, we use a similar architecture to RetICL but train using a supervised objective rather than RL. Specifically, we collect a dataset $\mathcal{D}$ of 20 prompts and resulting LLM outputs for each sample in the training and validation sets, where the examples in the prompts are randomly selected from the corpus. We feed samples $(x, e_1, \ldots, e_T) \in \mathcal{D}$ to an LSTM in the same way they are fed to RetICL. However, the LSTM classifier does not use a bilinear projection to define the policy $\pi(S_t, e)$, but rather uses a linear projection on LSTM hidden states to define an approximation of the $Q$-function $\hat{q}(S_t, e)$. We formalize this objective as
\begin{align*}
    & \hat{q}(S_{t-1}, e_t) = \mathbf{h}_{t}^T\mathbf{w}_q + b_q,\\
    & \ell = \sum_{t=1}^T (\hat{q}(S_{t-1}, e_t) - R^G)^2,
\end{align*}
where $\mathbf{w}_q \in \mathbb{R}^{d_h}$ and $b_q \in \mathbb{R}$ are learnable parameters, and $\ell$ is the loss for a single sample in $\mathcal{D}$. At inference time, we use the approximated $Q$-function to guide an example selection policy via greedy decoding, i.e., $e_{t+1} = \argmax_{e \in \mathcal{C}} \hat{q}(S_t, e)$.

We perform batched training on $\mathcal{D}$ and evaluate the \textit{classification accuracy} on the validation set after each epoch. We define this accuracy as the percentage of samples where the estimated value of the final example in the prompt correctly predicts if $\hat{y}$ will be correct or not, i.e., $\operatorname{sign}(\hat{q}(S_{T-1}, e_T)) = \operatorname{sign}(R^G)$. We train using the AdamW optimizer for 20 epochs with a learning rate of 1e-4, a weight decay of 0.01, a batch size of 256, and perform early stopping based on the classification accuracy on the validation set. We use the same training set, validation set, and training corpus as RetICL. We do not use soft prompts for S-BERT in this setting since we found that it hurts performance. We achieve classification accuracies of 85.54, 61.38, and 61.97 on TabMWP, GSM8K, and QASC, respectively. We note that these accuracies do not correlate with downstream performance when using the classifiers to guide example selection at inference time. This is likely because $\hat{q}$ is trained on a random policy but is used to guide a greedy policy at inference time, so there is an inherent difference between evaluations in these settings. Given these results, we can imagine a different inference time setting where the classifier is used to rank a list of randomly selected prompts rather than guiding a policy, but we leave this investigation for future work. Regardless, we believe these results further highlight the benefits of using RL algorithms for the example selection task.

\section{Scaling Number of In-Context Examples}
\label{sec:scaling-k}
\begin{figure}[h]
    \centering
    \includegraphics[width=.95\linewidth]{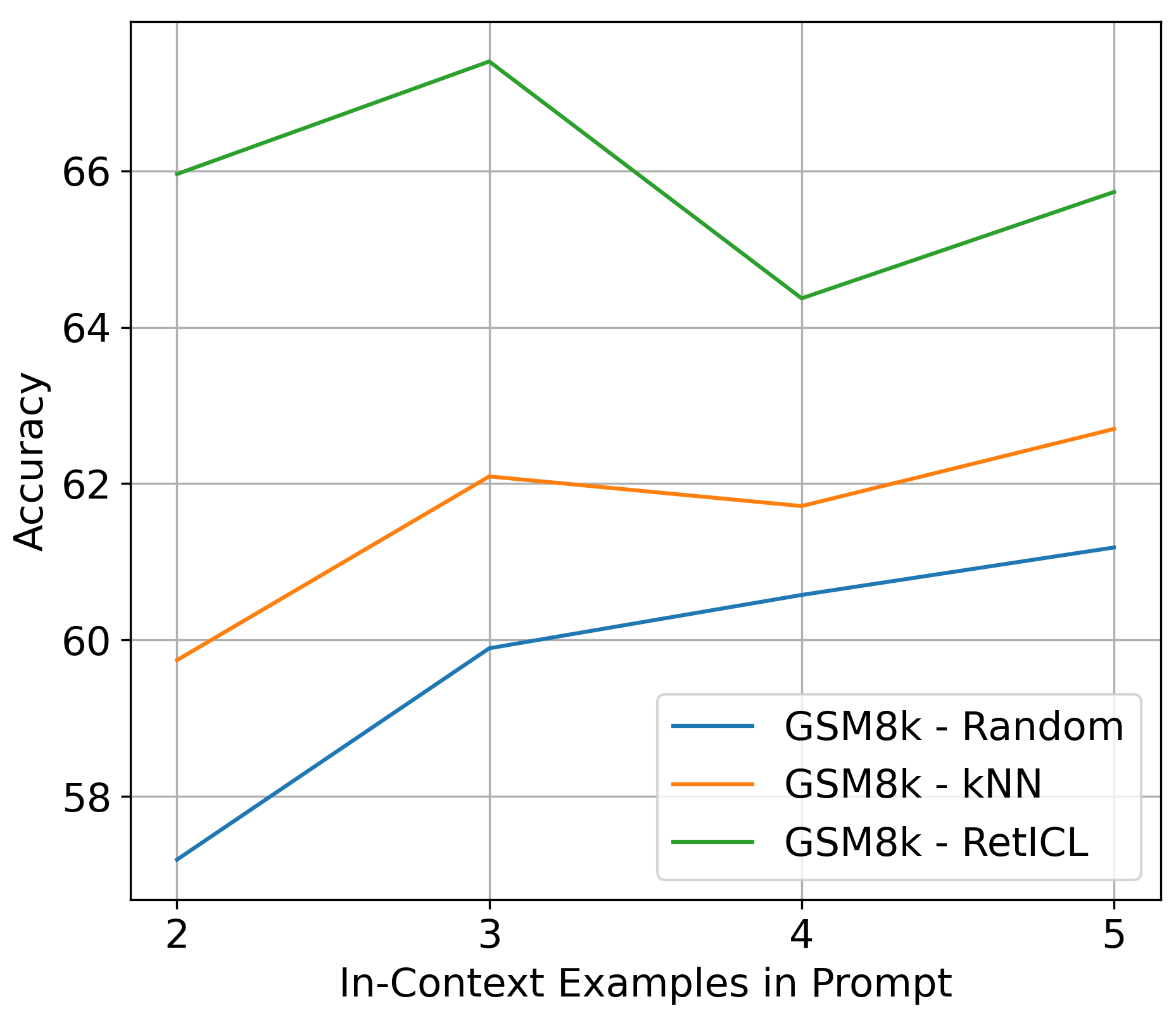}
    \caption{Accuracy as the number of in-context examples in the prompt increases, evaluated on GSM8K using prompting via Random, kNN, and RetICL.}
    \label{fig:scaling-T}
\end{figure}

We perform preliminary experiments on scaling $T$, the number of in-context examples used in each prompt. Specifically, with $T \in \{2,3,4,5\}$, we calculate the test set accuracy on GSM8K using prompting via Random, kNN, and RetICL (trained on 1,000 samples), and show the results in Figure \ref{fig:scaling-T}. We observe that while RetICL still performs better than baselines as $T$ increases, the gap between the methods narrow, particularly because RetICL performs roughly the same with $T=2$ as it does with higher values. While this result is unexpected, we note that finding an effective policy with a higher $T$ is a more complex task due to increased degrees of freedom. Therefore, we hypothesize that RetICL's inability to capitalize on more examples is due to training difficulties, and possibly because of our simpler PPO implementation. We leave further investigations of this phenomenon and potential training improvements for future work.

\section{Latent Space Analysis Details}
\label{sec:latent-space}

\begin{figure*}
    \centering
    \includegraphics[width=.45\linewidth]{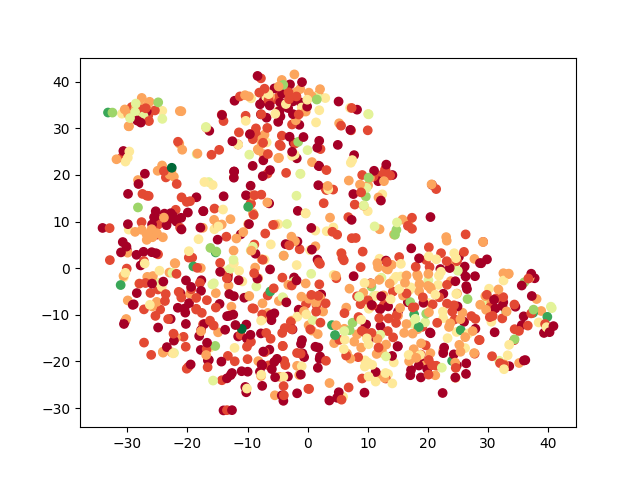}
    \includegraphics[width=.45\linewidth]{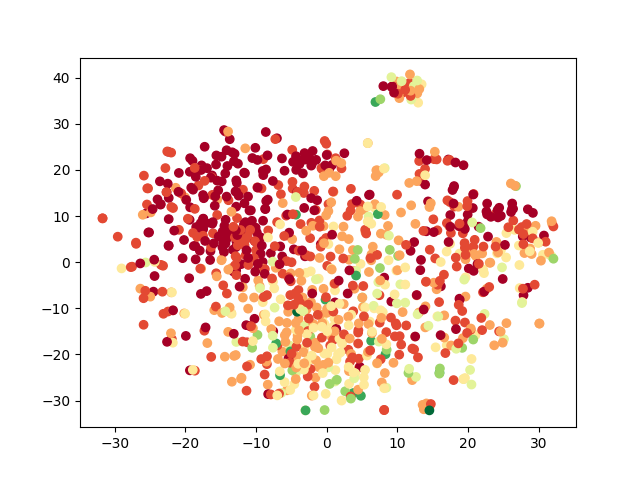}
    \includegraphics[width=.45\linewidth]{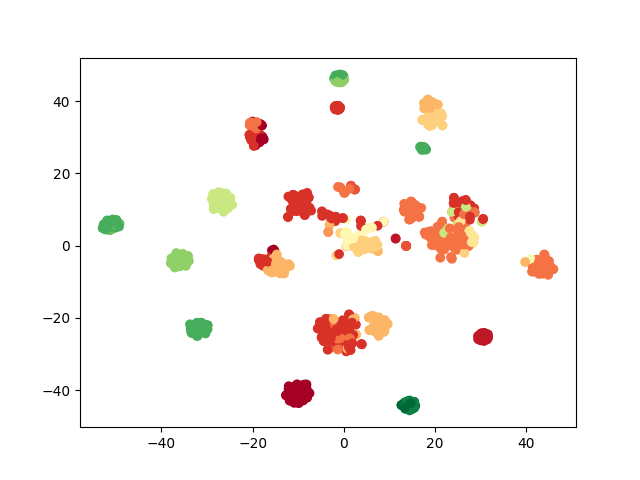}
    \includegraphics[width=.45\linewidth]{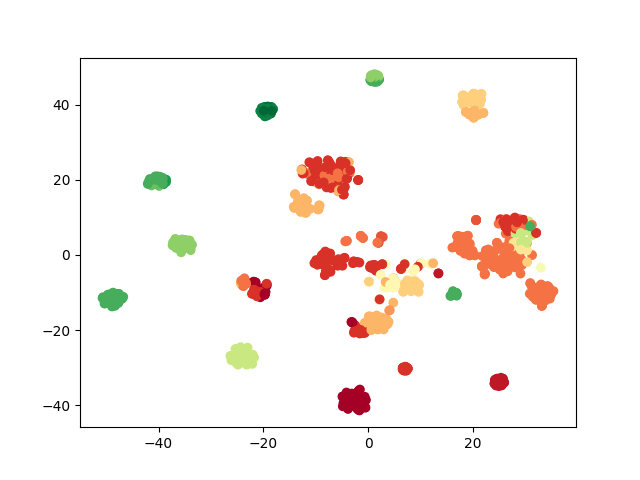}
    \caption{Example embedding visualizations. From left to right, top to bottom, GSM8K pre-trained embeddings, GSM8K RetICL embeddings, TabMWP pre-trained embeddings, and TabMWP RetICL embeddings. Points are colored based on the number of steps in an example's solution, with red being the least, green being the most, and yellow being in the middle.}
    \label{fig:latent_space}
\end{figure*}

We show the visualized example embeddings in latent space in Figure \ref{fig:latent_space}. For GSM8K, we see that RetICL groups examples based on the number of solution steps, whereas the pre-trained S-BERT embeddings do not.
We also see that clusters in the RetICL embeddings have been somewhat merged together from the pre-trained embeddings. This result can be interpreted by observing the pre-trained embeddings to be primarily clustered based on topic, e.g., problems about money and problems about time belong to separate clusters, since S-BERT embeddings reflect the semantic content of the examples. While local neighbors in the RetICL embedding space also tend to have similar topics, the clusters are less well-separated, which implies that both topic and the solution strategy, which is partly reflected in the length of solution steps, are used for example selection by RetICL.

For TabMWP, we see that the space looks very different from GSM8K, with many separate clusters being present in both the RetICL and pre-trained spaces, primarily based on the problem's template. For example, there is one cluster for asking yes/no questions about schedules, one for asking if someone has enough money to buy something, and one for asking what the mean of a set of numbers is. Since RetICL retains these clusters, we can infer that an example's template is key to example selection. This observation is also validated by kNN's high performance on this dataset, since the most semantically similar problems are always from the same cluster. While there are not many differences between the RetICL and pre-trained S-BERT embedding spaces, we observe that RetICL has pulled several clusters closer together. For example, it partially merges together problems that require finding the largest value and problems that require finding the smallest value from a set. This observation suggests that problems across the merged template clusters can be used interchangeably as examples, since their problems tend to have similar reasoning strategies.

\onecolumn
\section{Representative Example Selections}
\label{sec:examples}

\begin{table}[h]
    \centering
    \small
    \begin{tabularx}{\linewidth}{XX}
        \toprule
        \multicolumn{2}{c}{\textbf{Problem}}\\
        \multicolumn{2}{p{16cm}}{Marcus is half of Leo’s age and five years younger than Deanna. Deanna is 26. How old is Leo?}\\
        \midrule
        \multicolumn{2}{c}{\textbf{Gold Solution}}\\
        \multicolumn{2}{m{16cm}}{Marcus is 26 - 5 = 21 years old. Thus, Leo is 21 * 2 = 42 years old. Final Answer: 42}\\
        \midrule
        \multicolumn{1}{c}{\textbf{RetICL}} & \multicolumn{1}{c}{\textbf{kNN}}\\
        \midrule
        \multicolumn{2}{c}{\textbf{Selected Examples}}\\
        Problem: Katy, Wendi, and Carrie went to a bread-making party.  Katy brought three 5-pound bags of flour.  Wendi brought twice as much flour as Katy, but Carrie brought 5 pounds less than the amount of flour Wendi brought.  How much more flour, in ounces, did Carrie bring than Katy? Solution: Katy's three 5-pound bags of flour weigh 3*5=15 pounds. Wendi brought twice as much flour as Katy, or 15*2=30 pounds of flour. Carrie brought 5 pounds less flour than Wendi, or 30-5=25 pounds of flour. Thus, Carrie brought 25-15=10 pounds of flour more than Katy. In ounces, the 10 pound difference is 10*16=160 ounces. Final Answer: 160\newline\newline
        Problem: Vivian plays 10 Spotify songs every day. Her best friend Clara plays 2 fewer songs each day. If in June they didn't play any song during the weekends only, and there were 8 weekend days in June, what's the total number of songs they both listened to in that month? Solution: June has 30 days, so if they didn't play any songs on weekends, then they played 30 days - 8 days = 22 days. If Vivian played 10 songs each day, the total number of songs she played for the month is 10 songs/day * 22 days = 220 songs. Clara played 2 fewer songs, which is 10 songs/day - 2 songs/day = 8 songs/day. The total number of songs played by Clara for the whole month is 8 songs/day * 22 days = 176 songs. The total played by both is 220 songs + 176 songs = 396 songs Final Answer: 396
        & Problem: Leonard is 4 years younger than Nina who is half as old as Jerome. If the sum of their ages is 36, what is Leonard's age? Solution: Let x be the age of Nina. So, Leonard's age is x - 4. And Jerome's age is 2x. Since the sum of their ages is 36, then the equation is x + x - 4 + 2x = 36. By combining like terms, the equation becomes 4x = 40. So the value of x which represents the age of Nina is x = 40/4 = 10. Therefore, Leonard is 10 - 4 = 6 years old. Final Answer: 6 \newline\newline Problem: In 6 years, Carla will be 30 years old. The sum of the current ages of Carla and Louis is 55. How old is Louis now? Solution: Carla is currently 30-6=24 years old. Louis is 55-24=31 years old at present. Final Answer: 31 \\
        \midrule
        \multicolumn{2}{c}{\textbf{Generated Solutions}}\\
        Leo is twice Marcus' age, so Leo is 2*Marcus. Deanna is 5 years younger than Marcus, so Marcus is Deanna - 5. Deanna is 26, so Marcus is 26 - 5 = 21. Leo is 2*Marcus, so Leo is 2*21 = 42. Final Answer: 42 \color{green}\ding{51} & Leo is 26+5=31 years old. Marcus is 31/2=15.5 years old. Final Answer: 31 \color{red}\ding{55} \\
        \bottomrule
    \end{tabularx}
    \caption{Problem from GSM8K with examples and generated solutions from RetICL and kNN, where RetICL examples lead to a correct solution and kNN examples lead to an incorrect solution. A common feature between the current problem and the RetICL examples is that there are variables defined as offsets and multiplications of other variables. While the kNN examples are more semantically similar to the current problem, the required reasoning steps are very different. The generated RetICL solution more closely follows the style of the examples than the gold solution, breaking down the solution into more steps and using more verbal reasoning.}
    \label{tab:example_gsm8k_pos}
\end{table}

\begin{table}[]
    \centering
    \small
    \begin{tabularx}{\linewidth}{XX}
        \toprule
        \multicolumn{2}{c}{\textbf{Problem}}\\
        \multicolumn{2}{p{16cm}}{Table: [TITLE]: Pairs of shoes per store\newline
        Stem | Leaf \newline
        1 | 9\newline
        2 | 3, 9\newline
        3 | 2, 8\newline
        4 | 5\newline
        5 | 2\newline
        6 | 2, 3\newline
        Problem: Kristen counted the number of pairs of shoes for sale at each of the shoe stores in the mall. How many stores have at least 30 pairs of shoes but fewer than 40 pairs of shoes? (Unit: stores)}\\
        \midrule
        \multicolumn{2}{c}{\textbf{Gold Solution}}\\
        \multicolumn{2}{m{16cm}}{Count all the leaves in the row with stem 3. You counted 2 leaves, which are blue in the stem-and-leaf plot above. 2 stores have at least 30 pairs of shoes but fewer than 40 pairs of shoes. Final Answer: 2}\\
        \midrule
        \multicolumn{1}{c}{\textbf{RetICL}} & \multicolumn{1}{c}{\textbf{kNN}}\\
        \midrule
        \multicolumn{2}{c}{\textbf{Selected Examples}}\\
        Table: barrette | \$0.88\newline
        bottle of hand lotion | \$0.96\newline
        sewing kit | \$0.94\newline
        box of bandages | \$0.94\newline
        box of breath mints | \$0.80\newline
        Problem: How much money does Eve need to buy 6 bottles of hand lotion and a barrette? (Unit: \$)
        Solution: Find the cost of 6 bottles of hand lotion. \$0.96 × 6 = \$5.76 Now find the total cost. \$5.76 + \$0.88 = \$6.64 Eve needs \$6.64. Final Answer: 6.64 \newline
        \newline
        Table: [TITLE]: Rotten tomatoes per barrel\newline
        Stem | Leaf \newline
        2 | 0, 2, 6, 7\newline
        3 | 5, 6, 9\newline
        4 | 1, 5\newline
        5 | 4\newline
        6 | 1, 2, 9\newline
        7 | 2, 3, 5\newline
        Problem: The Clarksville Soup Company recorded the number of rotten tomatoes in each barrel it received. How many barrels had at least 70 rotten tomatoes but less than 80 rotten tomatoes? (Unit: barrels) Solution: Count all the leaves in the row with stem 7. You counted 3 leaves, which are blue in the stem-and-leaf plot above. 3 barrels had at least 70 rotten tomatoes but less than 80 rotten tomatoes. Final Answer: 3
        & Table: [TITLE]: Pairs of shoes per store\newline
        Stem | Leaf \newline
        3 | 5\newline
        4 | 2, 6\newline
        5 | 7\newline
        6 | 0, 4\newline
        7 | \newline
        8 | 4\newline
        9 | 0\newline
        Problem: Ron counted the number of pairs of shoes for sale at each of the shoe stores in the mall. How many stores have at least 50 pairs of shoes but fewer than 100 pairs of shoes? (Unit: stores)        Solution: Count all the leaves in the rows with stems 5, 6, 7, 8, and 9. You counted 5 leaves, which are blue in the stem-and-leaf plot above. 5 stores have at least 50 pairs of shoes but fewer than 100 pairs of shoes. Final Answer: 5 \newline
        \newline
        Table: [TITLE]: Pairs of shoes per store\newline
        Stem | Leaf\newline
        1 | 9\newline
        2 | 4\newline
        3 | 0, 0, 1\newline
        4 | 3\newline
        5 | 9\newline
        Problem: Emma counted the number of pairs of shoes for sale at each of the shoe stores in the mall. How many stores have at least 24 pairs of shoes but fewer than 46 pairs of shoes? (Unit: stores) Solution: Find the row with stem 2. Count all the leaves greater than or equal to 4. Count all the leaves in the row with stem 3. In the row with stem 4, count all the leaves less than 6. You counted 5 leaves, which are blue in the stem-and-leaf plots above. 5 stores have at least 24 pairs of shoes but fewer than 46 pairs of shoes. Final Answer: 5 \\
        \midrule
        \multicolumn{2}{c}{\textbf{Generated Solutions}}\\
        Count all the leaves in the row with stem 3. You counted 2 leaves, which are blue in the stem-and-leaf plot above. 2 stores have at least 30 pairs of shoes but fewer than 40 pairs of shoes. Final Answer: 2 \color{green}\ding{51} & Count all the leaves in the rows with stems 3 and 4. You counted 5 leaves, which are blue in the stem-and-leaf plots above. 5 stores have at least 30 pairs of shoes but fewer than 40 pairs of shoes. Final Answer: 5 \color{red}\ding{55} \\
        \bottomrule
    \end{tabularx}
    \caption{Problem from TabMWP with examples and generated solutions from RetICL and kNN, where RetICL examples lead to a correct solution and kNN examples lead to an incorrect solution. While the first RetICL example is seemingly unrelated to the current problem, the second example has the exact same reasoning steps as the current problem, enabling the LLM to copy the solution closely and get the correct answer. On the other hand, while both kNN examples match the current problem semantically, the reasoning steps are slightly different than what are required for the current problem, causing the LLM to get the incorrect answer by following the example reasoning steps too closely.}
    \label{tab:example_tabmwp_pos}
\end{table}

\end{document}